\documentclass[letterpaper]{article} % DO NOT CHANGE THIS
\usepackage{aaai2026}  % DO NOT CHANGE THIS
\usepackage{times}  % DO NOT CHANGE THIS
\usepackage{helvet}  % DO NOT CHANGE THIS
\usepackage{courier}  % DO NOT CHANGE THIS
\usepackage[hyphens]{url}  % DO NOT CHANGE THIS
\usepackage{graphicx} % DO NOT CHANGE THIS
\usepackage{natbib}  % DO NOT CHANGE THIS AND DO NOT ADD ANY OPTIONS TO IT
\usepackage{caption} % DO NOT CHANGE THIS AND DO NOT ADD ANY OPTIONS TO IT
\usepackage{algorithm}
\usepackage{newfloat}
\usepackage{listings}
\DeclareCaptionStyle{ruled}{labelfont=normalfont,labelsep=colon,strut=off} % DO NOT CHANGE THIS
\floatstyle{ruled}
\newfloat{listing}{tb}{lst}{}
\floatname{listing}{Listing}

\usepackage{graphicx}
\usepackage{amsmath, amssymb} % define this before the line numbering.
\usepackage{bm}
\usepackage{textcomp}
\usepackage{gensymb}
\usepackage{pifont}% http://ctan.org/pkg/pifont
\usepackage[T1]{fontenc}
\usepackage{multirow}
\usepackage{makecell}
\usepackage{xspace}
\usepackage{enumitem}
\usepackage{url}
\usepackage{enumerate}
\usepackage[noend]{algpseudocode} 
\usepackage{booktabs}

\makeatletter

\DeclareMathOperator*{\argmin}{arg\,min}
\DeclareRobustCommand\onedot{\futurelet\@let@token\@onedot}
\def\@onedot{\ifx\@let@token.\else.\null\fi\xspace}
\def\eg{\emph{e.g}\onedot} 
\def\ie{\emph{i.e}\onedot}

\def\ve{\boldsymbol}
\DeclareMathOperator{\Tr}{Tr}
\makeatother

\usepackage{tcolorbox}
\usepackage{fancybox}

\definecolor{global_grey}{RGB}{230,230,230}

\usepackage{xcolor} % 

\title{Bipartite Mode Matching for Vision Training Set Search \\ from a Hierarchical Data Server}

\author {
    Yue Yao\textsuperscript{\rm 1,{\dag}},
    Ruining Yang\textsuperscript{\rm 2},
    Tom Gedeon\textsuperscript{\rm 3}
}
\affiliations {
     \textsuperscript{\rm 1}Shandong University, China\\
    \textsuperscript{\rm 2}Northeastern University, United States\\
    \textsuperscript{\rm 3}Curtin University, Australia
}

\usepackage{bibentry}
\begin{document}

\maketitle
\let\thefootnote\relax\footnotetext{\dag\ Corresponding author: yaoyorke@gmail.com}

\begin{abstract}
We explore a situation in which the target domain is accessible, but real-time data annotation is not feasible. Instead, we would like to construct an alternative training set from a large-scale data server so that a competitive model can be obtained. For this problem, because the target domain usually exhibits distinct modes (\ie, semantic clusters representing data distribution), if the training set does not contain these target modes, the model performance would be compromised. While prior existing works improve algorithms iteratively, our research explores the often-overlooked potential of optimizing the structure of the data server. Inspired by the hierarchical nature of web search engines, we introduce a hierarchical data server, together with a bipartite mode matching algorithm (BMM) to align source and target modes. For each target mode, we look in the server data tree for the best mode match, which might be large or small in size. Through bipartite matching, we aim for all target modes to be optimally matched with source modes in a one-on-one fashion. Compared with existing training set search algorithms, we show that the matched server modes constitute training sets that have consistently smaller domain gaps with the target domain across object re-identification (re-ID) and detection tasks. Consequently, models trained on our searched training sets have higher accuracy than those trained otherwise. BMM allows data-centric unsupervised domain adaptation (UDA) orthogonal to existing model-centric UDA methods. By combining the BMM with existing UDA methods like pseudo-labeling, further improvement is observed. 
\end{abstract}

\begin{small}
\begin{links}
    \link{Code}{https://github.com/yorkeyao/BMM}
\end{links}
\end{small}

\section{Introduction}
\label{sec:intro}
% The effectiveness of deep learning models is largely dependent on the availability of labeled training data. To achieve high performance, these models typically require large-scale datasets with accurate annotations. In recent years, there has been a proliferation of datasets of various types, accompanied by a notable increase in their size. For instance, in detection datasets, the Pascal VOC~\cite{everingham2015pascal} dataset contains approximately 11.5 thousand images across 20 categories, while Object365~\cite{shao2019objects365} comprises about 638 thousand images in 365 categories. More recently, the COCO dataset~\cite{lin2014microsoft} includes around 123 thousand images and 80 categories. However, creating such datasets, especially those with manual labels, can be expensive. Thus, we consider an alternative, \ie, the training set search, where we aim to search for data from a large existing source pool.

The widespread adoption of machine learning models in real-world applications rely on the availability of extensive and well-annotated training datasets~\cite{lin2014microsoft,shao2019objects365,everingham2015pascal}. However, in certain domains with dataset bias, such as autonomous driving, medical imaging, or large-scale surveillance systems, real-time data annotation is often infeasible due to high costs, time constraints, and the need for domain expertise to deal with such data bias~\cite{torralba2011unbiased,fan2018unsupervised,zhong2019invariance,song2020learning,luo2020generalizing,bai2021person30k}. An alternative approach is to leverage large-scale, pre-existing data servers to search for transfer learning training sets that can still yield high-performing models. Yet, a critical challenge remains: the target domain often comprises distinct modes in filming scenes or object appearances, and if these modes are not well-represented in the training set, model performance is likely to suffer.

\begin{figure}[t]
\begin{center}
	\includegraphics[width=\linewidth]{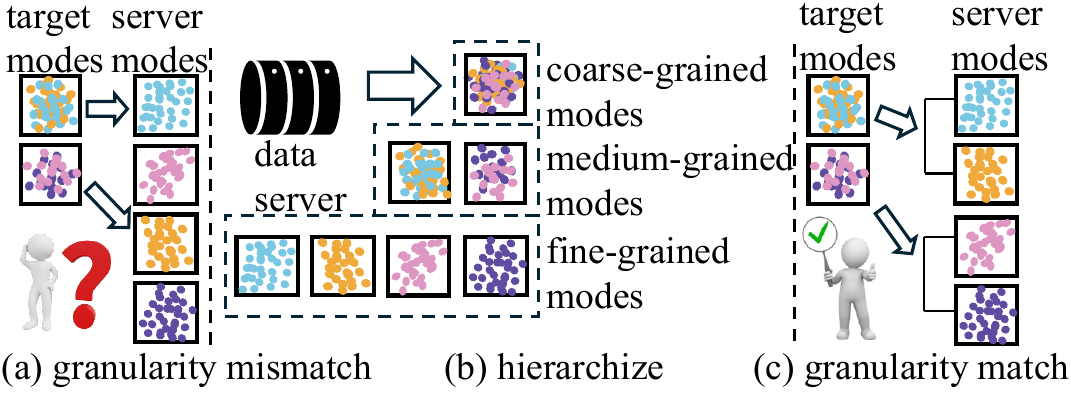}
\end{center}
\vspace{-1em}
\caption{Motivation illustration. Our research explores the
often-overlooked potential of optimizing the structure of the
data server. To explain, when aligning target modes with server modes, it is often challenging (a) due to granularity mismatches. In this paper, we propose a hierarchical server design (b) that allows target modes to match source modes at varying granularities, resulting in more effective and precise alignment (c).
% Inspired by the hierarchical structure of web search engines (\textbf{Left}), which employ algorithms like Pagelink~\cite{langville2006google} to organize and rank websites effectively, we adopt a similar approach for our data server. Our design introduces a hierarchical structure that categorizes data into different levels of granularity (\textbf{Right}). To construct a training set for a target domain from this server, we propose the source-target bipartite mode matching (BMM) method, which effectively retrieves the most suitable training data from the hierarchical data server.
 %We introduce the source-target bipartite mode matching (BMM) method for training set search from a large source pool. 
 }
\label{fig:intro}
% \vspace{-0.5em}
\end{figure}

We are motivated by the following considerations. First, while our goal is to align server and target modes, this alignment often faces a critical challenge of granularity mismatches. As illustrated in Fig.~\ref{fig:intro}, for example, if the target domain contains a mode labeled ``apple'', it would be inappropriate to align it only with specific modes like ``real apple'' or ``painted apples'', or with a broader category such as ``fruit''. A more meaningful match occurs when the semantics of the target mode and server mode align at a similar semantic level. For example, matching the target ``apple'' to a server mode that includes all forms of apples, whether real or depicted, is crucial for effective alignment. However, the semantics of modes are often difficult to control precisely, leading to inherent ambiguity in the alignment.

% Only when modes are aligned at the same semantic level can we achieve an accurate and relevant match.

% server modes and target modes, representing patterns clustered in the server and target domains, are likely to contain complex semantics. While they may correspond to a certain level of abstraction, the exact semantic hierarchy remains undetermined. This ambiguity in semantic granularity further complicates the alignment process. Specifically, the modes might be at either a high or low level of abstraction, and the alignment can only be effective when modes match at the same level of semantics. For example, in the case shown in Fig.~\ref{fig:intro}, if the target domain contains a mode in the semantic ``apple'', it would be inappropriate for it to match only specific modes like ``real apple'' or ``painted apples'' or larger semantic ``fruit''. A more meaningful match occurs only when the semantics of the target mode and the server mode align at a similar level—matching ``apple'' to a high-level semantic cluster that captures all forms of apples, whether real or depicted, is essential for effective alignment. Only when modes are aligned at the same semantic level can we achieve a relevant and accurate match.

Second, existing methods in this area, albeit just a few, rely on improving algorithms that optimize data search based on distribution difference~\cite{yao2023large,yan2020neural,cao2021scalable,tu2023bag} or model feedback~\cite{ghorbani2019data}. However, the potential of optimizing the structure of the data server itself is less studied. In our research, we address this gap by optimizing the architecture of data servers to better facilitate training set search. Specifically, we are inspired by web search engines, particularly the hierarchical organization used to retrieve and rank web pages efficiently. For example, Google's PageRank algorithm~\cite{langville2006google}, constructs a hierarchical structure of websites by evaluating the importance of web pages through link analysis. This hierarchical organization is essential for effective information retrieval and ensures that the relevant content is quickly accessible.

Given these considerations, our approach introduces a hierarchical data server that organizes server data into a multi-level structure, facilitating mode matching between the source and target domains. The hierarchical structure is specifically obtained by agglomerative clustering from bottom-up to build hierarchical semantics of modes. To leverage this hierarchical architecture, we also introduce bipartite mode matching (BMM) framework. BMM involves flat clustering for the target domain to generate target modes, based on their image features extracted using a model pre-trained on Imagenet~\cite{szegedy2016rethinking}. Afterwards, each target mode is linked to each server mode, with a feature-level distance between each pair of modes, specifically using the Fr\'{e}chet Inception Distance (FID)~\cite{heusel2017gans}. Based on these connection weights, we construct a bipartite graph composed of server modes and target modes as vertices, and their corresponding FIDs as edge values. By employing minimum weight bipartite matching, we select server clusters\footnote{We use ``cluster'' and ``mode'' interchangeably in this paper as modes are obtained using clustering.} to form an optimal training set distribution.

Experimentally, we show that the BMM, the joint use of a hierarchical data server and the bipartite matching, is superior than existing methods for training set search for object re-ID, and detection targets. 
% As a minor contribution, we also include a new benchmark on training set search for detection named Region100\footnote{https://github.com/yorkeyao/DataCV2024}, which include more difficulty on domain adaptive detection. 
We compare BMM with random selection and existing training set search techniques~\cite{yan2020neural,yao2022attribute} and observe that BMM leads to a consistently lower domain gap between the searched training set and target set, consequently providing the model with higher accuracy than competing methods. Additionally, we report that with further data pruning techniques, the refined training set can maintain or exceed the performance of the server data pool. Moreover, by combining the searched training set with existing unsupervised domain adaptation methods like pseudo-labelling, further improvement is observed.

\section{Related Work}
\label{sec:related_work}

\begin{figure*}[t]
% \vspace{-2em}
\centering
\includegraphics[width=0.96\linewidth]{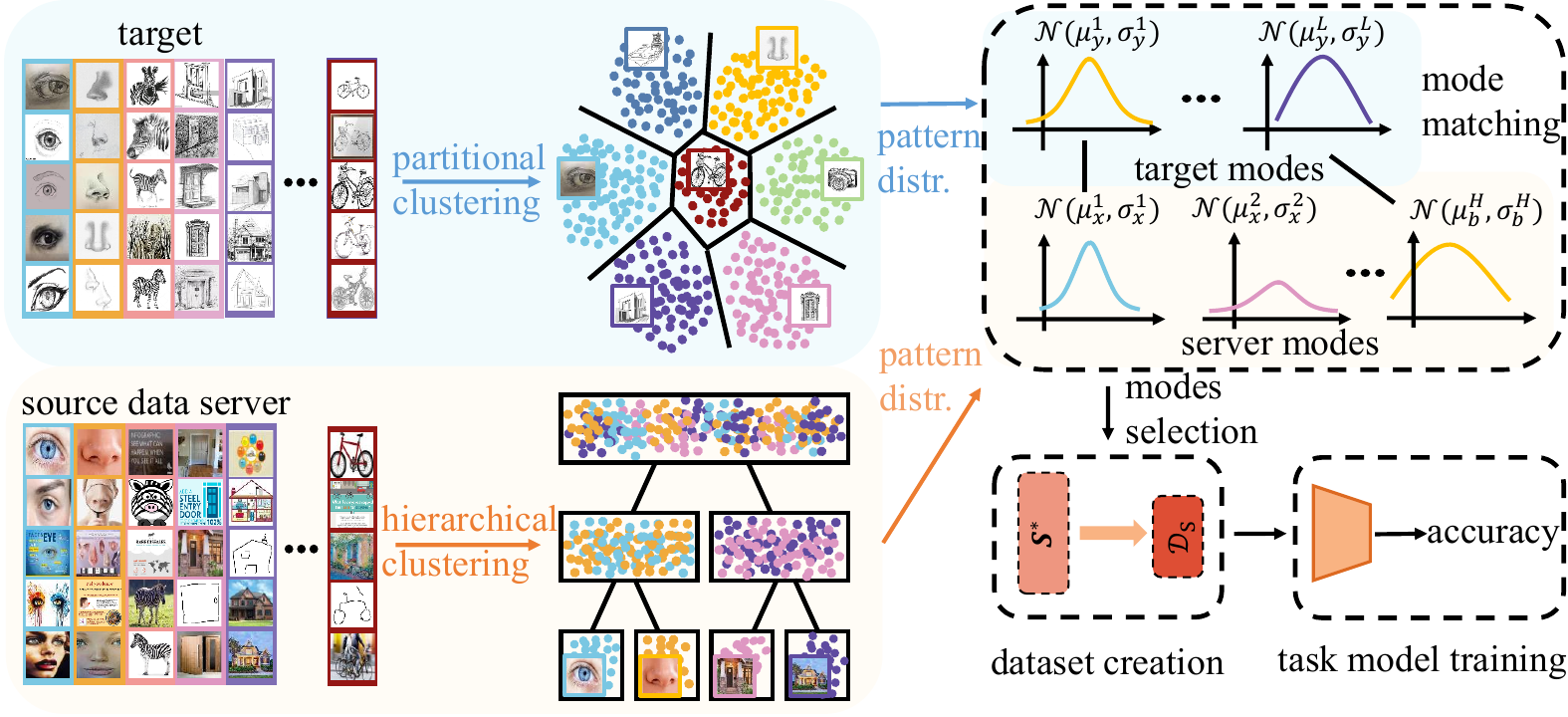}
\vspace{-1em}
\caption{ Workflow of BMM. 
% Specifically, we use the sketch in domainnet as the target and the rest of the other datasets in domainnet as the data server. 
\textbf{(Top left):} For a given target, we extract modes using flat clustering~\cite{lloyd1982least,macqueen1967some}. \textbf{(Bottom left):} For our data server, we extract modes using hierarchical clustering~\cite{mullner2011modern}. For modes existing in both target and the data server, we align them using the bipartite graph matching algorithm, \ie, Hungarian algorithm~\cite{kuhn1955hungarian,munkres1957algorithms}. For source modes aligning to the target, we select them to form our searched training set. The searched training set can be further pruned and then used for model training. 
}
\label{fig:BMM}
% \vspace{-0.5em}
\end{figure*}

% \subsection{Transfer learning}
% \textbf{Unsupervised domain adaptation} has been widely used in various fields, such as medical images analysis~\cite{ke2021chextransfer,mormont2018comparison}, language model~\cite{conneau2018senteval} and object detection~\cite{chen2017deeplab,dai2016r,girshick2014rich}. It utilizes knowledge gained from one task to improve the learning of another task, thereby reducing the impact of domain gaps. For most tasks in deep learning, the model's generalization ability is important. Thus, many researchers have attempted numerous learning algorithms to overcome domain gapes~\cite{torralba2011unbiased,perronnin2010improving,saenko2010adapting,deng2018image,lou2019embedding}. One strategy is domain adaptation to obtain a great performance model on a target domain by learning a transformation from the source domain to the target domain~\cite{hoffman2018cycada,lee2018diverse} or learning some common representation between two domains~\cite{bousmalis2016domain,tzeng2017adversarial,zhao2019learning}. In addition, domain generalization deals with domain gaps from different perspectives, including meta learning~\cite{balaji2018metareg,dou2019domain,li2018learning,zhang2020adaptive}, representation learning~\cite{khosla2012undoing,blanchard2011generalizing,gan2016learning,ghifary2016scatter,wang2020cross} and data augmentation. 
\textbf{Unsupervised domain adaptation} (UDA) has been widely applied in fields like medical image analysis~\cite{ke2021chextransfer,mormont2018comparison}, language modeling~\cite{conneau2018senteval}, and object detection~\cite{chen2017deeplab,dai2016r,girshick2014rich,yao2022attribute}, aiming to mitigate domain gaps by transferring knowledge from a labeled source domain to an unlabeled target domain. To enhance model generalization, various strategies have been explored~\cite{torralba2011unbiased,perronnin2010improving,saenko2010adapting,deng2018image,lou2019embedding}, including learning domain-invariant representations~\cite{bousmalis2016domain,tzeng2017adversarial,zhao2019learning} or performing source-to-target domain transformations~\cite{hoffman2018cycada,lee2018diverse,yao2019simulating,sunalice}. Moreover, domain generalization tackles this challenge from different angles, such as meta-learning~\cite{balaji2018metareg,dou2019domain,li2018learning,zhang2020adaptive}, representation learning~\cite{khosla2012undoing,blanchard2011generalizing,gan2016learning,ghifary2016scatter,wang2020cross}, and data augmentation.
The method introduced in this paper differs from these model-centric approaches. Our BMM framework uses data-centric methods \textbf{orthogonal to} these model-centric ones, and can achieve higher accuracy when jointly used.

% \subsection{Training set search}
% \textbf{Training set search from a data server}~\cite{yan2020neural, settles2009active, yao2023large, douze2024faiss} aims to search for training data from a large-scale data pool (server). For example, Xu \etal use a classifier with attention-based multi-scale feature extraction to select a small subset as training samples from a large pool of unlabeled data~\cite{xu2019positive}. Yan \etal do not directly search for training data but search for pre-trained data~\cite{yan2020neural,settles2009active}. They require unsupervised pre-trained experts to measure the domain gap. In contrast, we introduce a hierarchical data server and BMM algorithms eliminates these steps, streamlining the training set search approach and reducing the time needed. SnP~\cite{yao2023large} is the most close inspiring work to BMM. It achieves the training set by searching and pruning in the source data pool based on the user's budget. However, SnP is designed specifically for re-ID, but BMM is applicable to various tasks, including re-ID and object detection. 
\textbf{Training set search from a data server} focuses on retrieving effective training samples from large-scale data pools~\cite{yan2020neural, settles2009active, yao2023large, douze2024faiss}. While some methods~\cite{xu2019positive} use classifiers with multi-scale features to select subsets, others~\cite{yan2020neural, settles2009active} rely on pre-trained experts to assess domain gaps. The most relevant prior work, SnP~\cite{yao2023large}, searches and prunes data under budget constraints but is tailored for re-ID, whereas our method generalizes to broader tasks like object detection. Similarly, TL;DR~\cite{wang2023too} focuses on language processing and vehicle detection only. In contrast, we use a unified benchmark for broader evaluation across diverse tasks. In experiments, compared with these methods, we show that hierarchical structure and BMM leads to a consistently lower domain gap between the searched training set and target set, consequently yielding higher model accuracy than competing methods.

% \textbf{Dataset pruning} is the process of reducing training data without significantly impacting performance~\cite{yang2022dataset}. Clustering~\cite{agarwal2016second,agarwal2004approximating,feldman2020turning} and active learning~\cite{sener2017active} primarily rely on pre-defined criteria to calculate scores for each training example, followed by ranking the data and selecting examples with higher scores. Additionally, it is possible to synthesize a relatively smaller dataset with rich information through methods such as dataset distillation~\cite{wang2018dataset,such2020generative,sucholutsky2021soft,bohdal2020flexible,nguyen2020dataset,nguyen2021dataset,cazenavette2022dataset} or dataset condensation~\cite{zhao2020dataset,zhao2021dataset,zhao2023dataset,wang2022cafe,jin2021graph,jin2022condensing}. Howerver, the prerequisite of dataset pruning is that the training set data distribution aligns with the target domain~\cite{yao2023large}. However, in most cases, this assumption usually does not hold for searching from the server. BMM addresses this issue by conducting a search on the source data pool, directly obtaining a training set that aligns with the data distribution of the target domain. Subsequently, pruning the dataset can be safely performed to achieve an optimized training set with higher quality.

\section{Method}

Our goal is to build a labeled dataset from the source data pool for a given unlabeled target dataset, ensuring the differences in data biases compared to the target domain are as small as possible. This approach enables a model trained using the source dataset to perform effectively on the target dataset. In order to build the desired training dataset successfully, we propose the BMM framework.

\subsection{Problem Description}
We follow the search and pruning framework proposed in \citet{yao2023large}, with a special focus on the search part. The target dataset is defined as $\mathcal{D}_T=\{({\ve x}_i,y_i)\}_{i\in[m_t]}$, with $m_t$ representing the total number of image-label combinations in the target dataset, denoted by $[m_t]=\{1,2,\dots,m_t\}$. This set conforms to the distribution $p_T$, \ie,~$\mathcal{D}_T\sim p_T$.
% The source data pool $\mathcal{D}_S$ is constructed within the confines of a budget~$b$.

In order to create the training dataset $\mathcal{D}_S$, we establish a source data pool $\mathcal{S}$ comprising a variety of datasets or domains. This is expressed as $\mathcal{S}=\mathcal{D}_S^1 \bigcup \mathcal{D}_S^2 \dots \bigcup \mathcal{D}_S^K$, where each $\mathcal{D}_S^k$ for $k\in[K]$ denotes the $k$-th source dataset. Ideally, we hope to create a subset ${\mathbf S}^*$ from the data server $\mathcal{S}$ through a sampling process. Consider $h_{\mathbf S}$ as the model that is trained on any given dataset ${\mathbf S}$. The risk $h_{\mathbf S}$ of prediction on the test sample $\ve x$ which has the ground truth label $y$ is calculated as $\ell(h_{\mathbf S}({\ve x}), y)$. ${\mathbf S}^*$ is constructed with the aim of guaranteeing that the model $h_{{\mathbf S}^*}$ demonstrates the least risk on $\mathcal{D}_T$, \ie, 
\begin{equation}
{\mathbf S}^* = \argmin_{\mathbf S \in 2^\mathcal{S}} \mathbb{E}_{{\boldsymbol x},y \sim p_T}[\ell(h_{\mathbf S}({\ve x}), y)]. 
\label{eq:problem_define}
\end{equation}

\begin{algorithm}[t]  
  \caption{Bipartite Mode Matching for Training Set Search}  
  \begin{algorithmic}[1]
    \State \textbf{Input:} data server $\mathcal{S}$, target set $\mathcal{D}_T$, and number of source clusters $J$ and the number of target clusters $L$. 
    \State \textbf{Begin:}
    \State Balanced Cluster $(\mathcal{S}, J)\longrightarrow\{{\mathbf C}^1, \cdots, {\mathbf C}^J \}$ 
    \State $\{{\mathbf C}^1, \cdots, {\mathbf C}^J \} \longrightarrow \{{\mathbf S}^{1},\cdots,{\mathbf S}^{H} \}$     \Comment{Hier. clustering}
    \State Cluster $(\mathcal{D}_T, L)\longrightarrow\{{\mathbf T}^1, \cdots, {\mathbf T}^L \}$ 
    %\Comment{Cluster S}
    % \State ${\mathbf s}=\emptyset, \epsilon=\infty, {\mathbf S}^*=\emptyset$  %\Comment{Initialization}
    \State $V = \{{\mathbf S}^{1},\cdots,{\mathbf S}^{H}, {\mathbf T}^1, \cdots, {\mathbf T}^L \}$, $E=\emptyset$  \Comment{Graph init.}
    \For {$x$ in $1$ to $H$}
        \For {$y$ in $1$ to $L$}
            \State $edge({\mathbf S}^{x}, {\mathbf T}^{y}).value = \mbox{FID}({\mathbf S}^{x}, {\mathbf T}^{y}) $
            \State $E = E \mathop{\cup} edge({\mathbf S}^{x}, {\mathbf T}^{y})$
         \EndFor
    \EndFor
    \State $\sigma^*$ = Hungarian $(G(V, E))$ \Comment{bipartite graph matching}
    \State ${\mathbf S}^* = \emptyset$
    \For {$y$ in $1$ to $L$}
        \If {${\mathbf S}^{\sigma^*(y)} \notin {\mathbf S}^* $}
        \State ${\mathbf S}^* = {\mathbf S}^* \mathop{\cup} {\mathbf S}^{\sigma^*(y)}$
        \EndIf
    \EndFor
  \State \Return ${\mathbf S}^*$
  \end{algorithmic}  
  \label{algorithm:BMM}
\end{algorithm}

As analysed in \citet{yao2023large}, it is common for our target dataset $\mathcal{D}_T$ to have dataset bias (modes). In this case, if the training set cannot reflect them, the presence of such a difference introduces a hurdle to the efficacy of training, as models may not generalize effectively to real-world scenarios. In this paper, shown in Algorithm \ref{algorithm:BMM}, we propose the BMM, to search for a training set adapted to target bias.

\subsection{Hierarchical Data Server}

We are motivated by the notion that dataset bias can be represented by clustered modes~\cite{tu2023bag}. Given our data server, image features are extracted to create a sample space, and we then conduct clustering within this feature space. 

Specifically, we utilize a feature extractor $\bm{F}(\cdot)$ which converts an input image into a d-dimensional feature represented as $f \in \mathbb{R}^d$. Such a feature extractor, when pre-trained on Imagenet~\cite{szegedy2016rethinking}, is able to extract essential semantic features of $\mathcal{F}$ through efficiently condensing the image features into a refined representation. Formally, with $\mathcal{S}=\{({\ve x}_i,y_i)\}_{i\in[m_s]}$ representing the aggregate of $K$ datasets or domains and $m_s$ indicating the overall count of images after merging, we extract the features of all these images, donated as $\mathcal{F}_s = \{\bm{F}({\ve x}_i)\}_{i\in[m_s]}$. From these extracted features, a mode structure is established. Following this, we proposed to use balanced k-means clustering~\cite{lloyd1982least,macqueen1967some} to divide the feature space $\mathbb{R}^d$ into J clusters, \ie, $\{{\mathbf C}^{1},\cdots,{\mathbf C}^{J} \}$. Each cluster center among the J clusters is named as a ``mode'', since each center usually represents a distinct semantic content. Fig.~\ref{fig:BMM} shows representative images for each mode.  Thus formally, we aim to minimize the Sum of Squared Errors (SSE):
\begin{equation}
SSE = \sum_{k=1}^{J} \sum_{x_i \in C_k} \| x_i - \theta_k \|^2
\end{equation} which has the constraint that 
\begin{equation}
|C_k| = \frac{m_t}{K}, \quad \forall k
\end{equation}where $\theta_k$ means a cluster center and $SSE$ is the final loss. Subsequently, we merge clusters step by step according to the bottom-up hierarchical structure, \ie, agglomerative clustering~\cite{mullner2011modern}. Specifically, we start with each element in $\{{\mathbf C}^{1},\cdots,{\mathbf C}^{J} \}$, and pairs of closest clusters are merged as we move up the hierarchy. The process continues until all images are grouped into a single cluster, resulting in a dendrogram, \ie, a tree-like diagram illustrating the formation of clusters at each step. For all clusters with different sizes found during this process, we view them as distinct modes existing in the data server. By this process, we get $H$ modes, where finally we get $\{{\mathbf S}^{1},\cdots,{\mathbf S}^{H} \}$. Please note that $H$ is not a hyperparameter but a resulting number after the hierarchical merging of $J$ clusters. For example, in the case where the hierarchical structure is a full binary tree, the number of all nodes (modes) $H$ is approximately twice the number of all leaf nodes $J$, following $H = 2J-1$.
% Similarly,

A key of our method is the usage of balanced K-means rather than normal K-means. In K-means, mode sizes in the same hierarchical level may vary, as K-Means does not have constraints on this. That is, each mode at a given level does not necessarily contain the same number of images, which can lead to suboptimal mode matching in the following step when the mode size distribution follows a long-tail pattern. We replace K-Means with constrained K-Means, ensuring consistent mode sizes in a same level. Tab.~\ref{tab:constrained} shows that this simple yet task-aware refinement improves performance, highlighting the potential of further optimizing hierarchical structure construction.

\begin{figure*}[t]
% \vspace{-2em}
\centering
\includegraphics[width=\linewidth]{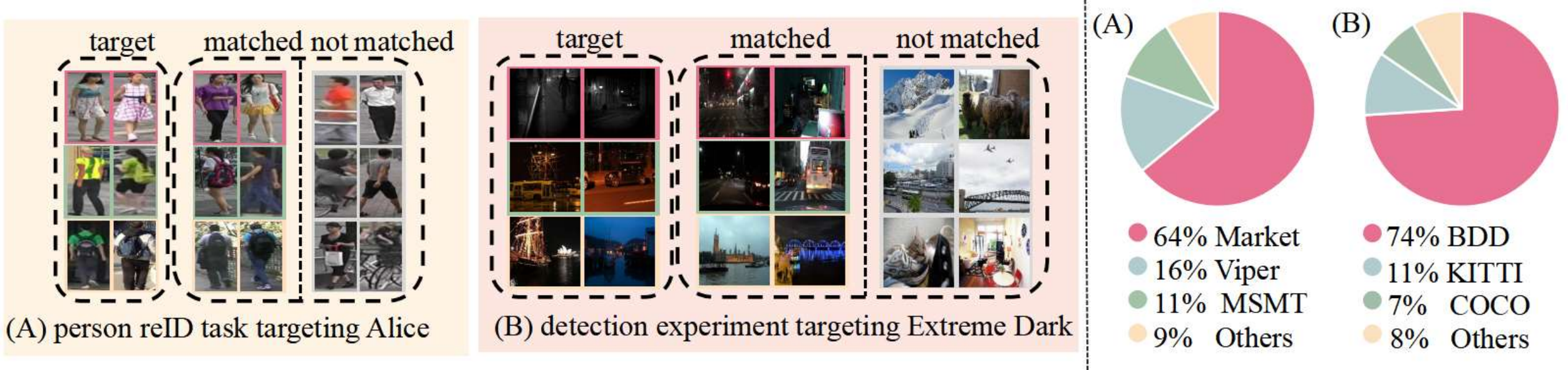}
\vspace{-2.3em}
\caption{ Mode matching examples (\textbf{Left}) and composition of the searched training set (\textbf{Right}). (\textbf{Left}): We show matched data server modes (images) resembling target modes (images). Four sub-figures (A) and (B) show mode matching examples for person re-ID, and vehicle detection respectively. For example, (B) shows a successful mode matching a dark environment. It is such successful mode matching that ensures the searched training set has a similar distribution to the target, and thus high training efficacy. (\textbf{Right:}) The pie charts on the right illustrate the proportions of the searched training set. 
% Sample images from target domain and source data pool \textbf{(Left)}, and the statistical composition of the searched training set \textbf{(Right)}. There are four sub-figure (A), (B), (C) and (D) on both left and right sides. Different subfigure show different tasks. In each sub-figure on the left, the left column shows image samples from the target domain. The right column is divided into two parts: the sample images that match the target domain searched through the BMM framework, and images that were not matched into the searched training set. The pie chart on the right illustrates the proportion of searched training set, with corresponding values detailed in the legend below each pie chart. Each subfigure corresponds one-to-one with the sample images displayed on the left: (A) is for the classification experiment with sketch as the target, (B) for the vehicle reID experiment with AliceVehicle as the target, (C) for the person reID task with AlicePerson as the target, and (D) for the detection experiment with Extreme Dark as the target.
}
\label{fig:pie_chart}
% \vspace{-0.5em}
\end{figure*}

% \begin{figure*}[t]
% \centering
%   % \begin{minipage}[c]{0.48\textwidth}
%     \includegraphics[width=0.7\linewidth]{images/para_analysis1.pdf}
%   % \end{minipage}\hfill
%   % \vspace{-1.8em}
%   % \begin{minipage}[c]{0.51\textwidth}
%     \caption{The comparison of source server flat clustering and source hierarchical clustering. We show the impact of the number of source and target clusters on model performance. 
%     % In both line graphs, the FID curves show the change in domain gap as the number of clusters increases. mAP curve shows the effect of the number of clusters on the accuracy of the model. 
%     The target cluster number equals 20 when varying source cluster numbers. Source cluster numbers equal 128 when varying target cluster numbers. Source flat clustering requires hyperparameter tuning for the ``sweet point'' while hierarchical clustering does not. 
%     % The left graph focuses on the source clusters, while the right graph pay attention to the target clusters.
%     }
%     \label{fig:para_analysis}
%   % \end{minipage}
%   \vspace{-1.5em}
% \end{figure*}

\begin{figure}[t]
% \vspace{-2em}
\centering
\includegraphics[width=\linewidth]{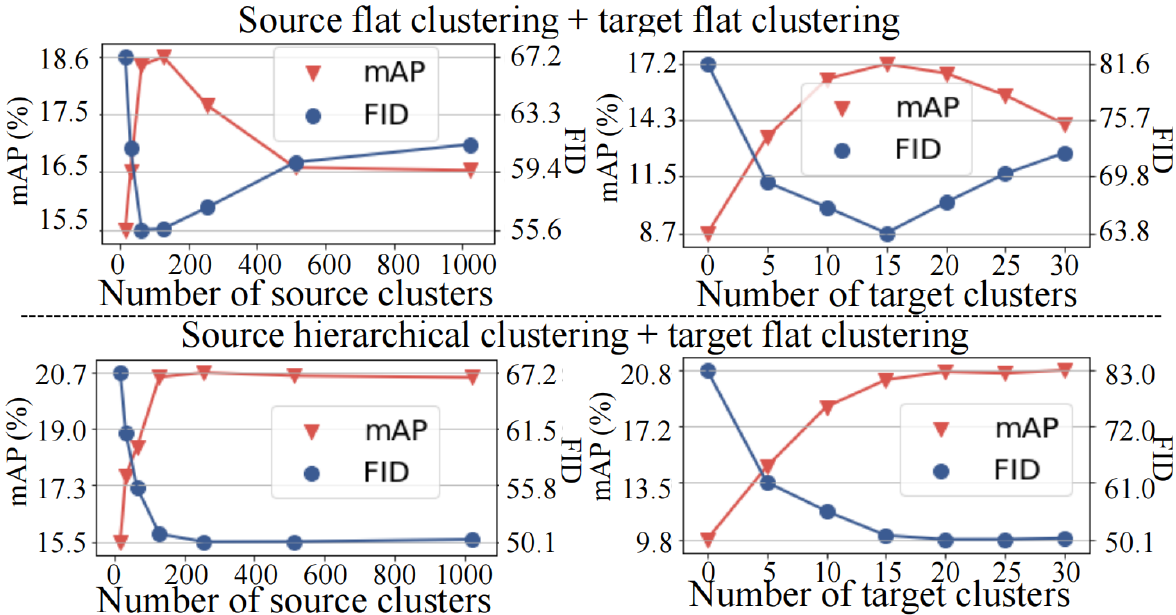}
\vspace{-1.6em}
\caption{The comparison of source server flat clustering and source hierarchical clustering. We show the impact of the number of source and target clusters on model performance. The target cluster number is fixed to 20 when varying source cluster numbers, and the source cluster number is fixed to 128 when varying target cluster numbers. Source flat clustering requires hyperparameter tuning for the ``sweet point'' (shown in \textbf{top left}) and \textbf{top right}, while hierarchical clustering does not (shown in \textbf{bottom left}). Furthermore, \textbf{bottom right} shows $L>20$ and \textbf{bottomleft} shows $>200$ source clusters (\ie, depth > 8) suffice for high accuracy.
}
\label{fig:para_analysis}
% \vspace{-0.5em}
\end{figure}

% \begin{figure*}[t]
% \vspace{-0.4em}
%   \begin{minipage}[c]{0.55\textwidth}
%     \centering
%     \includegraphics[width=\linewidth]{images/para_analysis1.pdf}
%   \end{minipage}
%   \hfill
%   \begin{minipage}[c]{0.45\textwidth}
%     \caption{The comparison of source server flat clustering and source hierarchical clustering. We show the impact of the number of source and target clusters on model performance. The target cluster number is fixed to 20 when varying source cluster numbers, and the source cluster number is fixed to 128 when varying target cluster numbers. Source flat clustering requires hyperparameter tuning for the ``sweet point'', while hierarchical clustering does not.}
%     \label{fig:para_analysis}
%   \end{minipage}
% \vspace{-2em}
% \end{figure*}

\subsection{Target-server Mode Matching}
\label{subsec:mmpm}
For each distinct mode existing in the target domain, our goal is to search for its corresponding similar modes in the source data pool. This can be solved by the algorithm designed for the assignment problem~\cite{kuhn1955hungarian}.

We extract image features and perform flat clustering on $\mathcal{D}_T$ to get $L$ clusters $\{{\mathbf T}^{1},\cdots,{\mathbf T}^{L} \}$. The flat clustering can help to find distinct modes in $\mathcal{D}_T$, and ensure all modes have enough dissimilarity with each other. 

The assignment problem can be instantiated as a complete undirected bipartite graph, where each edge is assigned a non-negative cost value. This is designed to precisely determine the cost associated with each link, thereby effectively addressing the assignment problem. In training set search, the undirected bipartite graph is represented as $\bm{G} = (\bm{V},\bm{E})$, with the vertex set $\bm{V}$ comprising the union of two non-intersecting sets, $\bm{X} = \{{\mathbf S}^{1},\cdots,{\mathbf S}^{H}\}$ (\ie, the sources modes) and $\bm{Y} = \{{\mathbf T}^{1},\cdots,{\mathbf T}^{L} \}$ (\ie, the target modes), such that $H > L$; In practice, we usually set the number of $H$ to two orders larger than $L$. The set E of edges contains every two-element subset $\{{\mathbf S}^{x},{\mathbf T}^{y} \}$, with ${\mathbf S}^{x} \in \bm{X}$ and ${\mathbf T}^{y} \in \bm{Y}$. The cost associated with edge $e = \{ {\mathbf S}^{x},  {\mathbf T}^{y} \}$, namely $c( {\mathbf S}^{x}, {\mathbf T}^{y})$, is determined using Fr\'{e}chet Inception Distance (FID)~\cite{heusel2017gans} between them, which is defined as:
\begin{equation}
% \begin{split}
\mbox{FID}(x, y) = \left \| \bm{\mu}_x - \bm{\mu}_y  \right \|^{2}_{2} + 
         \Tr(\bm{\Sigma}_x + \bm{\Sigma}_y -2 (\bm{\Sigma}_x \bm{\Sigma}_y)^{\frac{1}{2}}).
% \end{split}
\label{eq:fid}
\end{equation}
In Eq. \ref{eq:fid}, $\bm{\mu}_x \in \mathbb{R}^d$ and $\bm{\Sigma}_x \in \mathbb{R}^{d\times d}$ represent the mean and covariance matrix of the image descriptors for cluster ${\mathbf S}^{x}$, respectively. The mean and covariance matrix for cluster ${\mathbf T}^{y}$ are denoted by $\bm{\mu}_y$ and $\bm{\Sigma}_y$. The function $\Tr(.)$ denotes the trace of a square matrix, and $d$ refers to the dimension of the image descriptors. Given $\bm{G} = (\bm{V},\bm{E})$, to find a bipartite matching between these two sets $\bm{X}$ and $\bm{Y}$, we search for a permutation of L elements $\sigma \in \mathfrak{S}_L$ with the lowest overall cost of all matching:
\begin{equation}
% \begin{split}
\sigma^* = \argmin_{\sigma \in \mathfrak{S}_L} \sum_{i}^{L} \mbox{FID}({\mathbf T}^{i}, {\mathbf S}^{\sigma(i)}),
% \end{split}
\label{eq:matching}
\end{equation}
where $\mbox{FID}({\mathbf T}^{i}, {\mathbf S}^{\sigma(i)})$ refers to the matching cost between the original prediction and the perturbed prediction corresponding to index $i$ in $\{{\mathbf T}^{1},\cdots,{\mathbf T}^{L} \}$ and index $\sigma(i)$ in  $\{{\mathbf S}^{1},\cdots,{\mathbf S}^{H}\}$. This optimal pairing is efficiently calculated using the Hungarian algorithm following previous work~\cite{kuhn1955hungarian,munkres1957algorithms}. 

After this process, we can get our searched training set ${\mathbf S}^* = \{ {\mathbf S}^{\sigma(i)} \}_{i=1}^L$, means the combination all matched source clusters. Note that this process is also likely to have repeat selection of data. Thus, during the combination process, we delete those which previously appeared. After searching, we can use dataset pruning techniques to further reduce the training set size to meet our requirement~\cite{yao2023large}, resulting in $\mathcal{D}_S$ for model training.  

\section{Experiment}

% \subsection{A. Experimental Details}

% \textbf{BMM settings.} For a dataset denoted as a target domain, their unlabeled training sets are used as the target. 
% % For image feature extraction, we use IncepetionV3~\cite{szegedy2016rethinking} pretrained on ImageNet~\cite{deng2009imagenet}. 
 
% \textbf{Task model configuration.} Otherwise indicated, we use direct transfer models in this paper, where we train models on searched training sets and test on targets directly. Otherwise specified, % For classification, we use ResNet101~\cite{he2016deep}. 
% for object re-ID, we use ID-discriminative embedding (IDE)~\cite{zheng2016mars}. For detection, we use RetinaNet~\cite{lin2017focal}. 

% \textbf{Computation resources.} The experiments conducted in this paper were performed on a machine equipped with eight GPUs. Specifically, this machine is configured with eight NVIDIA GeForce RTX 3090 GPUs and two AMD EPYC 7343 Processors. On the software, our computational environment is based on PyTorch, utilizing version 1.12.1 with CUDA toolkit 11.6.  
\subsection{Server and Target Datasets}

For a dataset denoted as a target domain, their unlabeled training sets are used as the target. Otherwise indicated, we use direct transfer models in this paper, where we train models on searched training sets and test on targets directly. Unless otherwise specified, % For classification, we use ResNet101~\cite{he2016deep}. 
for object re-ID, we use ID-discriminative embedding (IDE)~\cite{zheng2016mars}. For detection, we use RetinaNet~\cite{lin2017focal}.

% \textbf{Classification.} We use DomainNet~\cite{peng2019moment}, which has 6 domains: Painting, Real, Infograph, Quickdraw, Sketch, and Clipart, for 345-class object classification. Each domain has its own training and test splits, with 123,973, 367,286, 71,106, 244,723, 7,345, and 34,019 images. Painting and Sketch are the targets, and the remaining 5 domains serve as the data server.

% \textbf{Classification.} We use domain generalization benchmark DomainNet~\cite{peng2019moment}, which consists of 6 domains: Painting, Real, Infograph, Quickdraw, Sketch and Clipart, where the tasks are 345-class object classification. Each domain has its training and test splits. We have 123,973, 367,286, 71,106, 244,723, 7,345, and 34,019 images for these domains respectively. In our experiment, we have Painting and Sketch domains serve as the target in turns, and the rest of the 5 domains serves as the data server.  

\textbf{Object re-ID.} 
We conducted experiments using both the person re-ID dataset and the vehicle re-ID dataset separately, reusing settings from \cite{yao2023large}.

\textbf{Vehicle Detection.} The data server is composed of seven datasets, including ADE20K~\cite{zhou2019semantic}, COCO~\cite{lin2014microsoft}, BDD~\cite{yu2020bdd100k}, CityScapes~\cite{cordts2016cityscapes}, DETRAC~\cite{wen2020ua,lyu2017ua}, Kitti~\cite{geiger2012we} and VOC~\cite{everingham2015pascal}. The data server has 176,491 images in total. %Seven detection datasets constitute a data server that contains in total of . 
We first use Exdark~\cite{Exdark}, which is captured only in low-light environments and has 7,363 images, as our target. We also use the Region100 benchmark as our target, which was used as the benchmark dataset in the 2nd CVPR DataCV Challenge.
%\footnote{https://sites.google.com/view/vdu-cvpr24/competition/}, 
The Region100 benchmark consists of footage captured by static cameras from 100 regions in the real world. For videos from each different region, the first 70\% is used for model training, while the remaining 30\% is designated for validation and testing. % More details to be shown in supplementary material. 

\textbf{Evaluation protocol.}  
% We use average top-1 accuracy for DomainNet classification. 
For object re-ID, we report mAP and CMC (``rank-1'' and ``rank-5''). In vehicle detection, mAP, mAP@50, and mAP@75 evaluate overall and threshold-specific detection accuracy at IoU of 0.5 and 0.75.
% In the DomainNet classification task, we present the average top-1 accuracy across all categories. In object re-ID task, we evaluate the system accuracy using the mean average precision (mAP) and cumulative match curve (CMC) metrics, such as ``rank-1'' and ``rank-5'' scores.  The ``rank-1'' indicates the probability of correctly identifying the true match at the top position, while ``rank-5'' represents the probability of having at least one true match in the top five rankings. In the vehicle detection task, we also use mAP as the main evaluation metric and combine it with its variants including mAP@50 and mAP@75. mAP provides a comprehensive performance metric by calculating the average precision at Intersection over Union (IoU) ranging from 0.5 to 0.95 in the average accuracy to evaluate the overall detection capability of the model. In comparison, mAP@50 and mAP@75 focus on specific IoU thresholds of 50\% and 75\%, respectively. These thresholds allow us to evaluate the model's performance under different accuracy requirements.

\subsection{Results and Discussion}
\label{sec:Quantitative_Evaluation}

\textbf{Working mechanism of BMM.} Shown in Fig.~\ref{fig:pie_chart}, in the process of BMM, we aim to align each distinct mode found in a target dataset with a corresponding one in the data server. 
This alignment ensures that both datasets contain similar modes, effectively reducing the bias difference that is the domain gap between the source data pool and target
\noindent\begin{minipage}[t]{\textwidth}
% \vspace{-1.1em}
\begin{minipage}[t]{0.21\textwidth}
\makeatletter\def\@captype{table}
\footnotesize
\centering
\vspace{-2.62em}
\begin{tabular}{c|cc}
    \Xhline{1.2pt}        Method    &   FID$\downarrow$   & mAP$\uparrow$    \\ 
\hline
Baseline   & 51.93 &   26.08  \\
Balanced  &  \textbf{50.48} & \textbf{28.16}  \\
\Xhline{1.2pt}
\end{tabular}
% \vspace{-0.8em}
\caption{Comparison to the balanced K-means. Settings are the same as Tab.~\ref{tab:snp_to_othermethods_reid} with 5\% pruning rate targeting AlicePerson. } %be-\\tween two feature extraction \\ model, \ie, Inception net \\ and ResNet. }
\label{tab:constrained}
\end{minipage}
\;
\begin{minipage}[t]{0.25\textwidth}
\makeatletter\def\@captype{table}
\footnotesize
\centering
\setlength{\tabcolsep}{0.2mm}
% \vspace{0.1em}
\begin{tabular}{c|ccc} 
\Xhline{1.2pt}        Method         & FID$\downarrow$ & SSIM$\uparrow$  & mAP$\uparrow$    \\ 
\hline
DM w. dup.   &  \textbf{51.07} & 15.85 &   20.14  \\
DM w./o. dup.  &  60.84 & \textbf{21.07} &   22.07  \\
BMM  &  51.93  &  20.45  &  \textbf{26.08}   \\
\Xhline{1.2pt}
\end{tabular}
\vspace{-0.9em}
\caption{Comparison between our method and direct match (DM). Settings are the same as Tab. ~\ref{tab:snp_to_othermethods_reid} with 5\% pruning rate targeting AlicePerson.  }
\label{tab:direct_match}
\end{minipage}
% \vspace{-0.8em}
\end{minipage}
domain. As proposed by \citet{yao2023large}, the closer the distribution of the source dataset is to that of the target, the more effective the training will be. Therefore, BMM facilitates the creation of a high-quality, target-specific training set by closely matching its distribution with the target. In many cases, it is very likely that we may not find a perfect semantic match for each mode in the server. However, our algorithm \textit{still strives to find the closest possible match} due to the minimization nature of the optimization objective shown in Eq.~\ref{eq:matching}. Thus, in these situations, we cab still reduce the domain gap and improve performance.

\begin{table*}[t]
\centering
\footnotesize 
% \vspace{-1em}
\setlength{\tabcolsep}{0.6mm}
% \resizebox{1\textwidth}{!}{
\begin{tabular}{l|l|l|ccc|ccc|ccc|ccc} 
\Xhline{1.2pt}
\multicolumn{3}{c|}{\multirow{3}{*}{Training data}}                  & \multicolumn{6}{c|}{Target domains in person re-ID}                                                                                                & \multicolumn{6}{c}{Target domains in vehicle re-ID}                                                                             \\ 
\cline{4-15}
\multicolumn{3}{c|}{}                                                & \multicolumn{3}{c|}{AlicePerson}                                                        & \multicolumn{3}{c|}{Market}                    & \multicolumn{3}{c|}{AliceVehicle}                                   & \multicolumn{3}{c}{VeRi}                        \\ 
\cline{4-15}
\multicolumn{3}{c|}{}                                                & FID$\downarrow$ & R1$\uparrow$ & \multicolumn{1}{l|}{mAP$\uparrow$} & FID$\downarrow$ & R1$\uparrow$ & mAP$\uparrow$ & FID$\downarrow$ & R1$\uparrow$ & \multicolumn{1}{l|}{mAP$\uparrow$} & FID$\downarrow$ & R1$\uparrow$ & mAP$\uparrow$  \\ 
\hline
\multicolumn{3}{c|}{data server}                                     & 81.67           & 38.96                            & 17.62                              & 37.53           & 55.55        & 30.62         & 43.95           & 30.47        & 14.64                              & 24.39           & 55.90        & 25.03          \\ 
\hline
\multicolumn{3}{c|}{Searched via BMM}                                & 51.08           & 50.63                            & 27.09                              & 27.68           & 60.57        & 36.05         & 21.27           & 47.84        & 27.24                              & 15.32           & 77.19        & 42.18          \\ 
\hline
\multirow{12}{*}{\rotatebox{90}{Pruning}} & \multirow{6}{*}{5\% IDs}  & Random        & 81.41           & 33.16                            & 14.49                              & 39.65           & 47.39        & 23.97         & 44.52           & 36.36        & 14.17                              & 25.27           & 70.38        & 30.44          \\
                         &                           & NDS~\cite{yan2020neural} & 61.01           & 44.63                            & 22.81                              & 31.63           & 49.17        & 24.77         & 32.48           & 41.32        & 17.77                              & 26.06           & 71.23        & 32.53          \\
                         &                           & SnP~\cite{yao2023large}  & 60.64           & 47.26                            & 25.45                              & 30.37           & 51.96        & 26.56         & 23.92           & 44.58        & 21.79                              & 18.09           & 72.05        & 36.01          \\
                         &                            & TL;DR~\cite{wang2023too} & 62.98           & 43.08                            & 21.95                              & 32.08           & 48.56        & 23.04        & 33.25           & 40.98        & 17.57                              & 25.98           & 71.23        & 32.53          \\
                         &                           & CCDR~\cite{chang2024classifier}  & 60.52           & 48.47                            & 25.04                              & 31.07           & 50.87        & 27.08         & 24.04           & 43.68       & 21.05                              & 17.89           & 71.58        & 35.85          \\
                         &                           & BMM           & \textbf{51.93}           & \textbf{49.28}                            & \textbf{26.08}                              & \textbf{27.05}           & \textbf{53.03}        & \textbf{28.39}         & \textbf{21.96}           & \textbf{45.08}        & \textbf{23.84}                              & \textbf{15.98}           & \textbf{72.69}        & \textbf{38.55 }         \\ 
\cline{2-15}
                         & \multirow{6}{*}{20\% IDs} & Random        & 79.33           & 38.10                            & 17.79                              & 38.63           & 53.15        & 28.39         & 43.90           & 40.89        & 18.13                              & 24.43           & 68.71        & 34.10          \\
                         &                           & NDS~\cite{yan2020neural} & 63.15           & 46.74                            & 22.65                              & 32.42           & 53.53        & 28.19         & 24.15           & 44.58        & 22.82                              & 18.74           & 71.04        & 38.07          \\
                         &                           & SnP~\cite{yao2023large}  & 61.87           & 47.20                            & 25.36                              & 30.58           & 57.14        & 33.09         & 23.47           & 46.07        & 25.24                              & 17.93           & 73.48        & 40.75          \\
                         &                            & TL;DR~\cite{wang2023too} & 65.12           & 45.48                            & 21.20                              & 33.87           & 51.08        & 26.48         & 25.08           & 43.05        & 21.08                              & 19.08           & 70.65        & 37.90          \\
                         &                           & CCDR~\cite{chang2024classifier} & 61.02           & 47.98                            & 25.08                              & 31.05          & 56.85        & 33.69         & 23.78           & 46.07        & 25.85                              & 17.30           & 73.78        & 41.07         \\
                         &                           & BMM           & \textbf{51.53}           & \textbf{49.68 }                           & \textbf{26.97}                              & \textbf{27.54}           & \textbf{60.49}        & \textbf{35.08 }        &\textbf{ 21.64}           & \textbf{47.34 }       & \textbf{26.18}                              & \textbf{15.72}           & \textbf{75.36}        & \textbf{42.05}          \\

\Xhline{1.2pt}
\end{tabular}
\vspace{-0.5em}
\caption{Comparing different methods in training data search: random sampling, greedy sampling, greedy search, and proposed BMM, in \textbf{object re-ID tasks}. We set the pruning rate as $5\%$ and $20\%$ of the total source IDs. We use four targets: AlicePerson, Market, AliceVehicle and VeRi. The task model is IDE~\cite{zheng2016mars}. FID, rank-1 (\%), and mAP (\%) are reported. 
}
% }
\label{tab:snp_to_othermethods_reid}
% \vspace{-0.5em}
\end{table*}

\begin{table*}[t]
\centering
\footnotesize
% \vspace{-1em}
\setlength{\tabcolsep}{0.9mm}
\begin{tabular}{l|l|l|cccc|cccc} 
\Xhline{1.2pt}
\multicolumn{3}{l|}{Training data} & \multicolumn{4}{c|}{ExDark} & \multicolumn{4}{c}{Region 100} \\
\cline{4-11}
\multicolumn{3}{l|}{} & FID$\downarrow$ & mAP$\uparrow$ & mAP@50$\uparrow$ & mAP@75$\uparrow$ & FID$\downarrow$ & mAP$\uparrow$ & mAP@50$\uparrow$ & mAP@75$\uparrow$ \\ 
\hline
\multicolumn{3}{l|}{data server} & 104.98 & 40.43 & 79.56 & 36.71 & 251.47 & 19.65 & 41.38 & 17.25 \\ 
\hline
\multicolumn{3}{l|}{Searched via BMM} & 56.95 & 45.36 & 82.71 & 42.31 & 140.82 & 23.52 & 50.21 & 18.25 \\ 
\hline
\multirow{6}{*}{\rotatebox{90}{Pruning}} & \multirow{6}{*}{5\% Imgs} & Random & 105.74 & 23.50 & 56.13 & 14.49 & 251.90 & 11.38 & 25.30 & 9.14 \\
& & NDS~\cite{yan2020neural} & 64.35 & 30.34 & 65.34 & 22.18 & 165.31 & 12.62 & 25.34 & 9.00 \\
& & SnP~\cite{yao2023large} & 59.78 & 32.15 & 67.18 & 25.69 & 153.82 & 15.07 & 34.65 & 11.34 \\
& & TL;DR~\cite{wang2023too} & 66.25 & 28.32 & 65.11 & 22.07 & 162.34 & 12.87 & 25.08 &  9.08\\
& & CCDR~\cite{chang2024classifier} & 57.98 & 32.96 & 67.67 & 25.03 & 142.48 & 22.25 & 44.49 & \textbf{19.68} \\
& & BMM & \textbf{56.34} & \textbf{34.83} & \textbf{76.57} & \textbf{25.83} & \textbf{140.07} & \textbf{23.08} & \textbf{46.34} & 18.08 \\
\Xhline{1.2pt}
\end{tabular}
\vspace{-0.5em}
\caption{Comparing different methods in training data search: random sampling, greedy sampling, greedy search, and proposed BMM, in the \textbf{vehicle detection task}. We set the pruning ratio as $5\%$. The task model is RetinaNet~\cite{lin2017focal}. Two targets are used: Exdark and Region100. We report FID, mAP (\%), mAP@50 (\%) and mAP@75 (\%) respectively. 
% We observe \textbf{Region100 is more difficult than Exdark}.
}
\vspace{-0.5em}
\label{tab:snp_to_othermethods_detect}
\end{table*}

\textbf{Benefits of using hierarchical clustering on data server.} As mentioned in Fig.~\ref{fig:intro}, we use hierarchical clustering rather than flat clustering to minimize the impact of mode granularity (\eg, the semantic granularity of each cluster found by clustering) during matching processes. Our objective is to enable each target mode to match with the most similar modes in semantics from the data server, irrespective of source server mode size, which can be controlled by the hyperparameter: number of source clusters.

As shown in Fig.~\ref{fig:para_analysis}, if we use flat clustering on the source, we will need to carefully adjust both the number of source clusters and the number of target clusters to ensure competent matching between them, thereby reaching the ``sweet point'' as shown in Fig.~\ref{fig:para_analysis} upper part. Such a ``sweet point'' shows that at such a specific source and target mode size, source modes and target modes can be relatively competently matched, thus resulting in a competent training set searched. However, finding such a ``sweet point'' requires significant effort in hyperparameter tuning and is not practical in implementation.

In comparison, as shown in Fig.~\ref{fig:para_analysis} bottom, as long as the number of source clusters is not considerably small, performance will not be compromised. The reason is that if we use hierarchical clustering on the source, we prepare hierarchical modes with different levels of semantics (mode sizes) in our data server in preparation for being matched to the target. Thus source mode size will not be an influential factor of performance anymore.

Further, compared with hierarchical clustering on the source, flat clustering results in higher FID and lower mAP, showing poor matching between target and source modes.

\begin{figure*}[t]
% \vspace{-2em}
\centering
\includegraphics[width=0.9\linewidth]{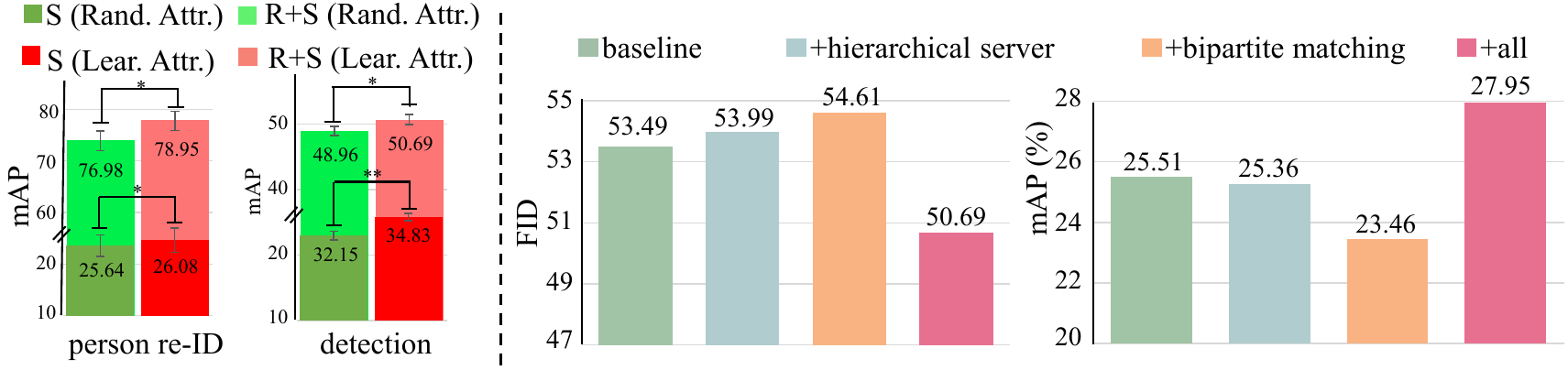}
\vspace{-0.5em}
\caption{\textbf{Left:} Joint usage of the training set search and existing UDA (\ie, pseudo-labelling) methods. MMT~\cite{ge2020mutual} and AT~\cite{li2022cross} are used for re-ID and detection respectively. \textbf{Right:} Ablation study of the hierarchical server and BMM framework (target Aliceperson). Starting from the baseline, we gradually add the hierarchical design and the mode matching module, observing consistent performance improvements.
%     % Specifically, $*$ means statistically significant (\ie $0.01 < p$-value $< 0.05$) and $**$ denotes statistically very significant (\ie $p$-value $< 0.01)$. Pseudo labels obtained by the UDA methods can further augment our searched training set to achieve higher accuracy.
}
\label{fig:error_bar}
% \vspace{-0.5em}
\end{figure*}

\textbf{Necessity of using flat clustering on target data.} In the BMM framework, our goal is to match each distinct target mode with a corresponding source mode. Using flat clustering allows us to efficiently identify distinct modes, ensuring minimal correlation between them. If hierarchical clustering were applied to the target data, higher-level modes would strongly correlate with lower-level ones, complicating the task of selecting appropriate modes from redundant ones for constructing the data server. Thus, flat clustering is necessary for effective mode matching.

\textbf{The rationale for using Hungarian matching instead of a simpler greedy direct matching.}
 Compared with direct matching (DM), the Hungarian matching algorithm prevents multiple target modes from matching the \textbf{same} source mode, which, though rare, occurs in practice (\eg, 4 out of 20 cases when targeting AlicerPerson). As shown in Tab.~\ref{tab:direct_match}, direct match allowing duplicate matches keeps FID low but reduces data diversity (measured in SSIM averaged across datasets), harming performance. Simply deleting these duplicates leaves some target modes unmatched, degrading FID and model performance. Bipartite graph matching ensures one-to-one mode assignment, balancing diversity and domain gap, thus improving results.

% \textbf{Hierarchical clustering builds on clusters rather than directly on images.} 
% % Our approach uses hierarchical clustering based on existing clusters instead of directly on images. In our design, we first carry out flat clustering and then apply hierarchical clustering to the clusters obtained. 
% Although performing hierarchical clustering on images directly might bring higher cluster accuracy, this design may have very high time complexity. For example, when performing hierarchical clustering on $\{{\mathbf C}^1, \cdots, {\mathbf C}^J \}$. It takes $\mathcal{O}(J^2logJ)$~\cite{murtagh2012algorithms}. This time complexity is not acceptable, especially with a very large number $J$, let alone the number of images directly. In our experiments with detection datasets, we have a very large data server that contains 176,491 images. A $\mathcal{O}(J^2logJ)$ algorithm directly on 176,491 images is too time-consuming. To save time and also maintain hierarchical structure, we first perform flat clustering and then perform hierarchical clustering on the clusters obtained. 

% However,  To be shown in our experiment, we further tried to applied hierarchical clustering on target, 

\textbf{Computing time complexity analysis.} Considering $L$ target clusters and $J$ source clusters obtained, and $J >> L$. The overall time complexity of our BMM framework is $\mathcal{O}(J^3)$, including $\mathcal{O}(L^2)$ for target flat clustering, $\mathcal{O}(J^2logJ)$ for source hierarchical clustering, and for matching. Please note that the hierarchical process is analogous to an offline training phase and only needs to be done once. For different target domains, we do not need to reconstruct the hierarchical data server each time. Our mode-matching algorithm, which is applied for each target domain, has a comparatively lower time complexity ($\mathcal{O}(logJ*J*L)$). This ensures that the computational cost remains manageable during deployment, as the expensive hierarchical structure construction is a one-time process that only happens in preprocessing. 

\textbf{BMM framework \emph{vs.} random sample and existing training set search algorithms.} We have this analysis in 
% Table \ref{tab:snp_to_othermethods_cls}, 
Table \ref{tab:snp_to_othermethods_reid} and Table \ref{tab:snp_to_othermethods_detect}. To make a fair comparison and eliminate the influence of dataset size, we searched for subsets of equal size via uniform sampling. For the comparison method, SnP was originally designed for re-ID only, we extended their method to detection~\cite{yao2023large}. TL;DR was originally designed for language only, we extended their method to vision tasks and excluded TL;DR's data generation, and evaluated only its data search component~\cite{wang2023too}. CCDR was designed for object detection, we extended it to person re-identification~\cite{chang2024classifier}.  Experiments on different tasks have shown a significant reduction in the domain gap between the searched dataset compared to the source server and the target domain, and an increase in the accuracy of the model.

For person re-ID on AlicePerson with a 5\% pruning ratio, random sampling yields an FID of 81.41, rank-1 accuracy of 33.16\%, and mAP of 14.49\%, whereas BMM reduces the FID by 29 and improves rank-1 accuracy and mAP by 16\% and 12\%, respectively. Similar gains are observed on vehicle re-ID. Compared with NDS~\cite{yan2020neural}, SnP~\cite{yao2023large}, TL;DR~\cite{wang2023too}, and CCDR~\cite{chang2024classifier}, BMM achieves the best domain alignment and detection performance, reaching an FID of 56.26, mAP of 42.23\%, and mAP@50 of 81.69\%. The improvement over the greedy search in SnP is statistically significant and stable (Fig.~\ref{fig:error_bar}). Pseudo labels from UDA further enhance performance.

\textbf{Joint usage of the training set search and existing UDA methods yields higher accuracy.} The training set search methods are orthogonal to existing UDA methods, \eg, pseudo labels obtained by the latter can further augment our searched training set to achieve higher accuracy. As shown in Fig. \ref{fig:error_bar} left, our method, when jointly used together with pseudo-labeling methods Mutual Mean-Teaching (MMT)~\cite{ge2020mutual} and Adaptive Teacher (AT)~\cite{li2022cross}, yields much higher accuracy for re-ID (targeting Market), and detection, respectively. For example, the accuracy of using MMT increases from 26.08\% to 78.95\% compared to only using the searched training set for direct transfer. When joint training with MMT, our method increases from 76.98\% to 78.95\% compared to greedy search.

\textbf{Ablation study.} We conduct an ablation study of the proposed modules, as shown in Fig.~\ref{fig:error_bar} right. Our baseline uses flat clustering in the data server and a greedy search strategy to select clusters~\cite{yao2023large}. Replacing flat clustering with hierarchical clustering alone does not lead to noticeable improvement. Similarly, using the proposed mode-matching module in place of greedy search, while keeping flat clustering, also yields no performance gain. These results are expected—our system is designed for hierarchical clustering and mode matching to work in tandem. Only when both modules are used together do we observe significant improvements over the baseline, confirming the effectiveness of our joint design. The composition of searched training sets is shown in Fig.~\ref{fig:pie_chart}.

\textbf{Analysis of hyperparameters.} 
Fig.~\ref{fig:para_analysis} (bottom) shows the relationship between the number of clusters in the source ($J$) and target ($L$) domain clustering, mAP, and FID metrics. As $J$ and $L$ increase, the domain gap (FID) and mAP improve. In the source domain, accuracy stabilizes at $J=128$. In the target domain, accuracy plateaus at $L=10$, and the FID reaches its lowest value of 9.8 at $L=20$, with mAP at 83.0\%. 
% We set $J=128$ for the source and $L=20$ for the target to optimize performance. 
As long as the cluster numbers are not too small, performance remains stable.

\section{Conclusion}

In this paper, we focus on the training set search problem from a data server, for object re-ID and detection tasks, with a specific focus on the structure of the source data server. We propose a hierarchical data server and BMM framework, to make the searched training set have a similar distribution with the target, such as styles and class distributions. We show that the matched source modes constitute training sets that have consistently better mode matching and smaller domain gap with the target domain. Experiments show that the BMM outperforms existing training set search methods. Furthermore, we analyze various components in the BMM system and find them to be stable under various data servers and targets, and hyperparameters.

\section*{Acknowledgments}
This work was supported in part by the Key Research and Development Program of Shandong Province China (2025CXGC010901), the Shandong Province Overseas Young Talents Program, the ARC Discovery Project (DP210102801), Oracle Cloud credits, and related resources provided by Oracle for Research.

\bibliography{aaai2026}
\clearpage
\label{sec:reproducibility-checklist}
% \input{checklist/ReproducibilityChecklist}

\clearpage
% \appendix
% \input{sections/supplemental_material}

\end{document}